\documentclass[12pt]{article}
\usepackage{graphicx}
\usepackage{url}
\usepackage[
top    = 0.5in,
bottom = 0.5in,
left   = 1in,
right  = 1in]{geometry}
\usepackage{caption}
\usepackage{subcaption}
\usepackage{array}
\usepackage{amsmath,amsthm,amssymb}
\usepackage{booktabs}
\usepackage{wrapfig}
\begin{document}
\title{  \textbf{Traffic Prediction in Cellular Networks using Graph Neural Networks} }
\author{Maryam Khalid }

\date{}
\maketitle

\section{Introduction}

Cellular networks are ubiquitous entities  that provide major means of communications all over the world. A cellular network is composed of  three main components, service provider, cellular users and base stations that connect the cellular users to the service provider. The density of base stations and cellular users is extremely high and is much more significant than the backhaul  setup. Thus, the base station and cellular users constitute major part of  cellular network whose study and evaluation is extremely critical as the number of users  and their requirements  are changing rapidly and new challenges are emerging.\\

One major challenge in cellular networks is dynamic change in number of users and their usage of telecommunication service which results in overloading at certain base stations. One class of solution to deal with this overloading issue is deployment of drones \cite{uav} that can act as temporary base stations and offload the traffic from the overloaded base station. There are two main challenges in development of this solution. Firstly, the drone is expected to be present around the base station where an overload would occur in future thus requiring a prediction of traffic overload. Secondly, drones are highly constrained in their resources and can only fly for few minutes. If the affected base station is really far, drones can never reach there. This requires initial placement of drones in sectors where overloading can occur thus again requiring a traffic forecast but at a different spatial scale.\\

It must be noted that the spatial extend of region that the problem poses and the extremely limited power resources available to the drone pose a great challenge which is hard to overcome without deploying the drones in strategic positions to reduce the time to fly to required high-demand zone. Moreover, since drones fly at finite speed, it is important that a predictive solution that can forecast traffic surges is adopted so that drones are available to offload the overload before it actually happens. Both these goals require an analysis and forecast of cellular network traffic which is the main goal of this project.

\section{Problem Statement}

 The focus of this project is on cellular traffic datasets  where traffic values are changing in both spatial and temporal domain.  It has been shown in literature that traffic values are correlated in both space and time i.e if one base station has high traffic, its neighboring base stations would also experience high traffic, Residential areas would experience high traffic at nigh during off work hours and commercial areas would experience high traffic during working hours. These are the observations that can be made from the data. However, there is a lot more too that makes these spatiotemporal dynamics of traffic data very complex to model. Since, deriving models is baffling, the only  resort  is learning it from data and this is where machine learning or more precisely neural networks come into the picture. \\
 
 Traditional solutions develop models \cite{model_based} for traffic prediction based on time series or derive solutions based on statistical time-series analysis. However, these methods are unable to capture the dependencies between different base stations which can prove useful in achieving the above mentioned goals.\\

In this project, we aim to exploit the dependencies between different base station to achieve two goals. The first  is to utilize the given data set to learn and estimate the traffic values in certain spatial  regions given information about others. Secondly, given the current and past traffic data we want to predict the future traffic pattern.\\

The relation between different entities or base stations is best captures by a graph. Once a graph is constructed, the base station(BS) characteristics which are node features and the structure of the graph can both be utilized to learn traffic values at other BS (Nodes) where data is not available. The relationship between traffic values and the node features and graph structure is pretty complex to be derived analytically. Therefore, in this project we exploit Neural networks to learn this relationship. \\

Please note that the traffic data we are considering in this project is time-stamped base station locations along with their traffic values which  is not  in the form of a graph. Even if we treat all BS as nodes, the definition of an 'edge' or dependency between BS is not implicit which makes the  process of converting this data into meaningful graph non-trivial. Therefore, the first task in this project is defining the notion of graph which is discussed in the next section.\\

Once a graph is formed, the next task is to feed it to a neural network. For that we need an embedding for the network that captures the structure of the graph and the important features at the same time. Please note that the dataset we are looking at is at the scale of a city and therefore, we require our NN to be designed in such a way that its agnostic of the number of input parameters. For that we utilize Convolutional neural networks where the parameters to be learned are equal to the size of the convolution filter . More details on this are provided in upcoming section.\\

After the construction of graph and its embedding, in the first part of the project we  pose the problem in spatial domain only i.e  given labels about certain nodes, we learn the labels for others that are not provided. In this particular application, the label is chosen to be high or low traffic where High traffic can raise a flag and guide the drones to move closer to those nodes.
In the second part of the project, we introduce \textit{time} and pose the problem in spatio-temporal domain where given certain traffic patterns, the machine learning model predicts the traffic patterns in the next few minutes.

\subsection{Graph Construction}
Before explaining the construction of graph it is important to process the data and take a look at it. The data considered in this project has two independent variables : location and time. These two variables lead to two output objectives : estimating traffic at some BS location whose value is not available and predicting traffic in future time.\\

 For this project, I am considering the dataset of cellular users\cite{data} for different base stations at a time resolution of few minutes for the city of Milan. The total area is divided into a grid of 10000 squares. The activity for incoming sms, outgoing sms, incoming call, outgoing call and internet usage is seperately recorded. All the activity in a given grid square is recorded as one value by averaging over the area of grid square. Furthermore, the temporal data is aggregated over 10-minutes time windows.\\
 
 Some of the data processing that I had to do involved mapping the BS ids to their locations and converting the GPS locations given in WGS84 (EPSG:4326) standard to Cartesian coordinates for ease of visualization and computation of euclidean distance. Moreover, the data is not provided in temporal sequence so I generated 'snapshots' of traffic data for unique time stamps and then visualized them. The video of visualization can be viewed at \cite{videolink}.
 \\
 It is not possible to explain the patterns in traffic distribution over space and time through visualization of raw plots only. However, just to showcase existence of a dependency and motivate the use of graph to capture them, I have presented some snapshots in Figure (1).
 
 \begin{figure}
 	
 	\includegraphics[scale=0.6]{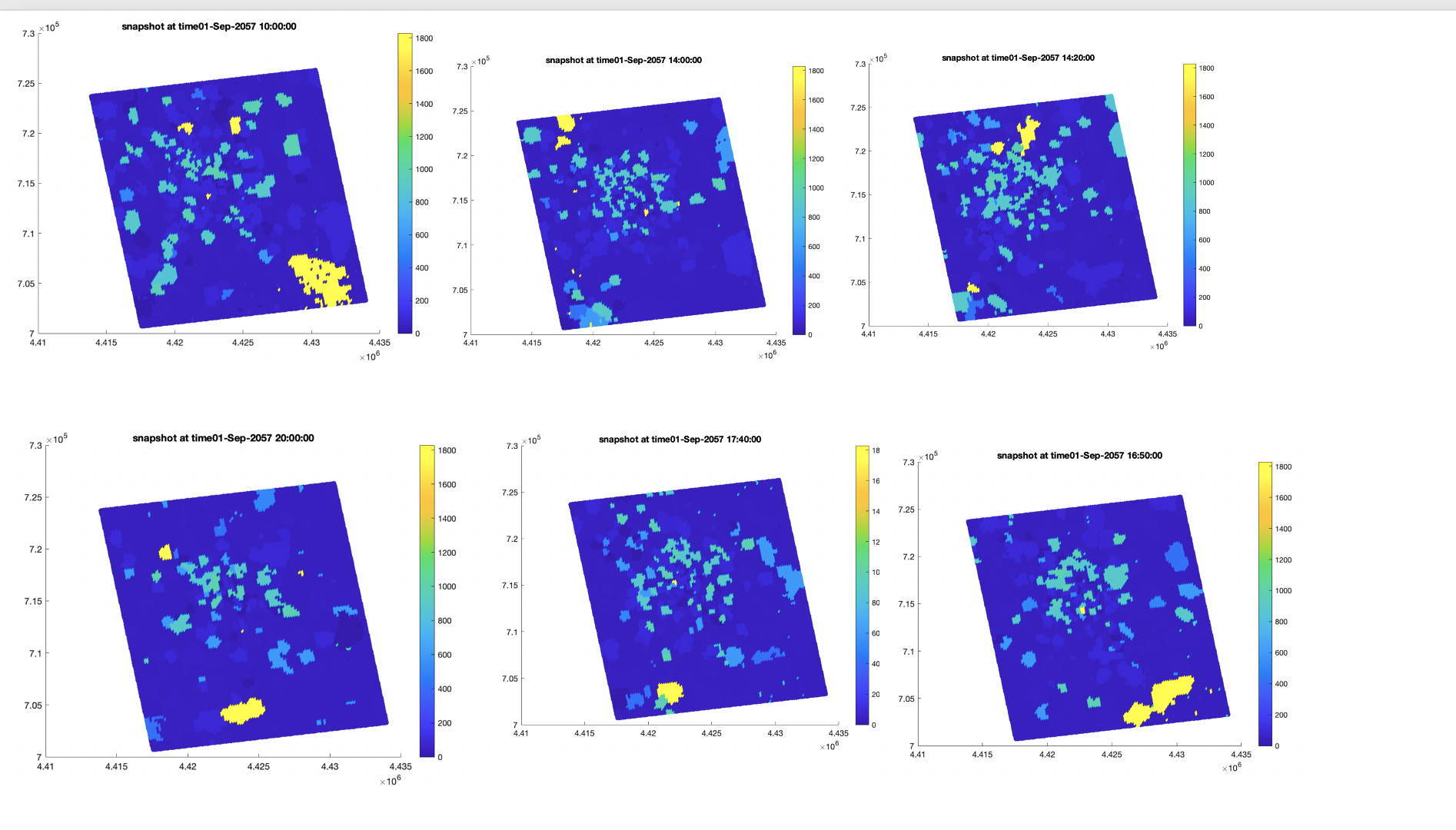}
 	\caption{Traffic pattern in city of Milan at different times. }
 \end{figure}

The x-axis and y-axis represent the 2-D locations on Cartesian plane. The color bar shows the traffic value and as we move from left to right the time is increasing. So the rightmost snapshot is that of the morning and we see that the traffic value is very high in left bottom of plots in the morning and in the evening (last plot) indicating that it might be a residential area. Furthermore,we can observe that there are several squares in the center of grid that share similar values. This figure is not sufficient to make any solid conclusions but at least it justifies the intuition that when traffic demand is high at one point it is high in the surrounding spatial region too therefore we see small clusters of yellow and light blue. 
Also some regions might be correlated with far away regions because people commute from residential area to commercial areas on weekdays. \\

Now for constructing the graph, the first metric I used is spatial proximity which is defined as follows,

\begin{equation}
	f(x_i,x_j)=\begin{cases}
		1, & \text{if $||x_i-x_j||_2<\epsilon$}.\\
		0, & \text{otherwise}.
	\end{cases}
\end{equation}

where $x_i$ and $x_j$ are the nodes of  a graph. For this metric, I define the node as the BS.

\begin{figure}[h!]
	\includegraphics[scale=0.65]{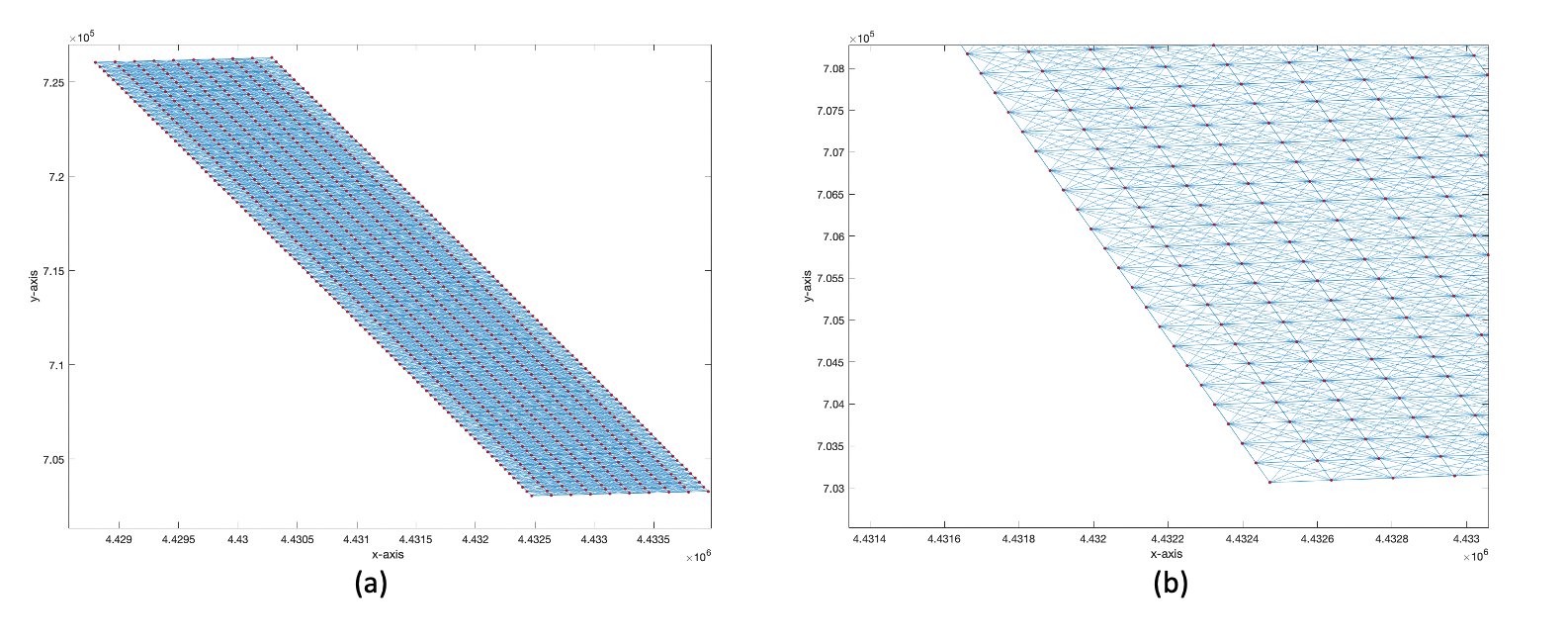}
	\caption{(a). Original Graph with non-weighted adjacency ,matrix (b). Zoomed in view to visualize edges}
\end{figure}

in the above graph, all nodes within the radius $\epsilon$ are given equal weight which is not very ideal if large radius is considered. For that weighted adjacency matrix is defined as, 

\begin{equation}
	f(x_i,x_j)=\begin{cases}
		exp(-||x_i-x_j||_2^2/\sigma^2), & \text{if $exp(-||x_i-x_j||_2^2/\sigma^2)>\epsilon$} \;\; \&\;\; i \neq j \\
		0, & \text{otherwise}.
	\end{cases}
\end{equation}

where $\sigma$ controls the maximum magnitude of weights .

\subsection{Graph Embedding}
In order to represent the graph in a vector form that can then be processed through a neural network to generate estimated and predicted traffic values at different nodes and times, we need graph embedding. Please note that in this section I am working with static graphs or snapshots only. \\

Consider a graph G whose adjacency and degree matrix is represented by A and D. There are many methods to embed the graph in a low-dimensional space. Methods like Laplacian eigenmap project the nodes in the reduced subspace spanned by leading eigenvectors of the laplacian of graph G. The method is straightforward and therefore I am skipping the details here for the time being. For this method, I consulted this paper \cite{lap}. The limitation of this method is that it is computationally very expensive for large networks like the one we are considering here in this work as it requires computation of eigenvectors of laplacian. Conpared to this method, graph convolution network (GCN) is much more computationally efficient. The parameters are shared across all nodes and the number of parameters only depend on the size of convolution filter. \\

In GCN, there are several variants of the filters that are used for convolution with input features and producing a low-dimensional output. For a graph with N nodes, let X represent the input feature matrix of size NxF where F is feature size. Each layer of the GCN can be represented as a function of a matrix L and input H. The input to the first layer is X. The matrix L is what defines a rule for the filter. For instance, a representation of the node as aggregate of its features can be represented by $L=D^{-1}\hat{A}$ where $\hat{A}=A+I$ to include the node's own contribution. Other variants include $L=A$, $L=D^{-0.5}AD^{-0.5}$  etc. Finally an activation is applied and the final form of each GCN layer can written as :
\begin{equation}
	f(L,H_i) = ReLu(L*H_i*W_i)
\end{equation}
Where $W_i$ are weights of layer $i$.
Currently, I have used  $L_1=D^{-1}\hat{A}$  and $L_2=D^{-0.5}AD^{-0.5}$  and in my implementation and I call them method 1 and method 2 respectively.
The results for a comparison between laplacian and GCN for 1000 nodes and random weight initialization are provided in figures below.\\

\begin{figure}
	\includegraphics[scale=0.4]{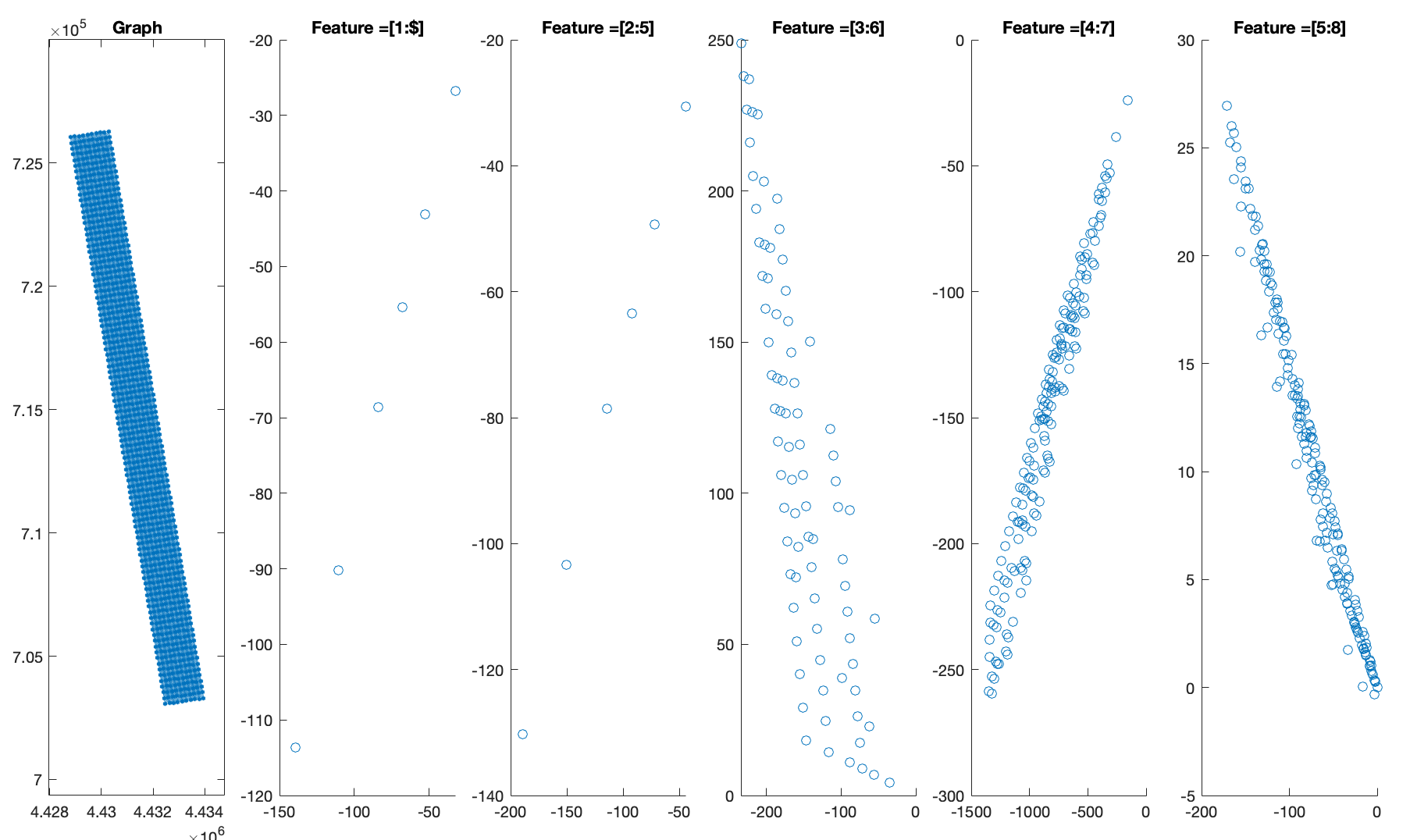}
	\caption{Method 1 embedding with different combinations of features chosen from a set of 8 features}
\end{figure}

\begin{figure}
	\includegraphics[scale=0.4]{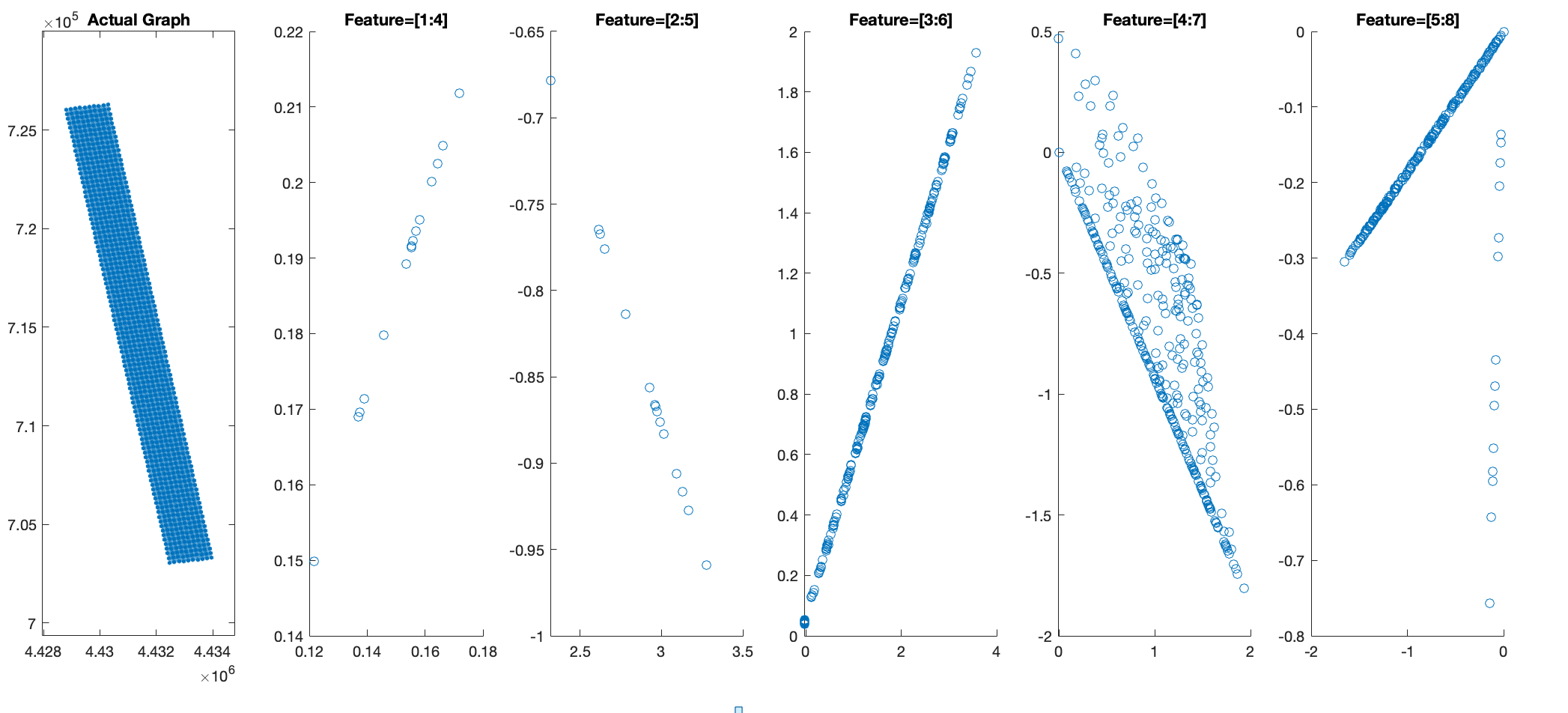}
	\caption{Method 2 embedding with different combinations of features chosen from a set of 8 features}
\end{figure}

For the embedding provided by method 1, it is hard to make any conclusions because the graph only shows the spatial dependence. In order to take a further look I evaluated the embeddings with different subsets of features space which are presented in figure (3).  It can be observed that certain features like the first 2 plots are mapping most of the nodes to same locations because of similar text and call values. However, the last three  plots shows that features 7-8 that are representing internet traffic are providing some better mapping. \\

In method 2 as well, internet traffic is providing more distinguishable embedding. Comparing method 1 with 2, the latter gives better embedding in terms of mapping nodes to more distinguished locations. 
Finally, the laplacian embedding presented in Figure (5) is computed by using all 8 features. It can be observed that this embedding is a combination of  embeddings obtained by different feature combinations in 1 and 2. for instance, this triangle can be formed by adding the 2nd,3rd and 4th subfigure in figure (5). The laplacian embedding seems reasonable but is computationally very expensive as eigenvectors of the laplacian have to be computed.

\begin{figure}[h!]
	\includegraphics[scale=0.4]{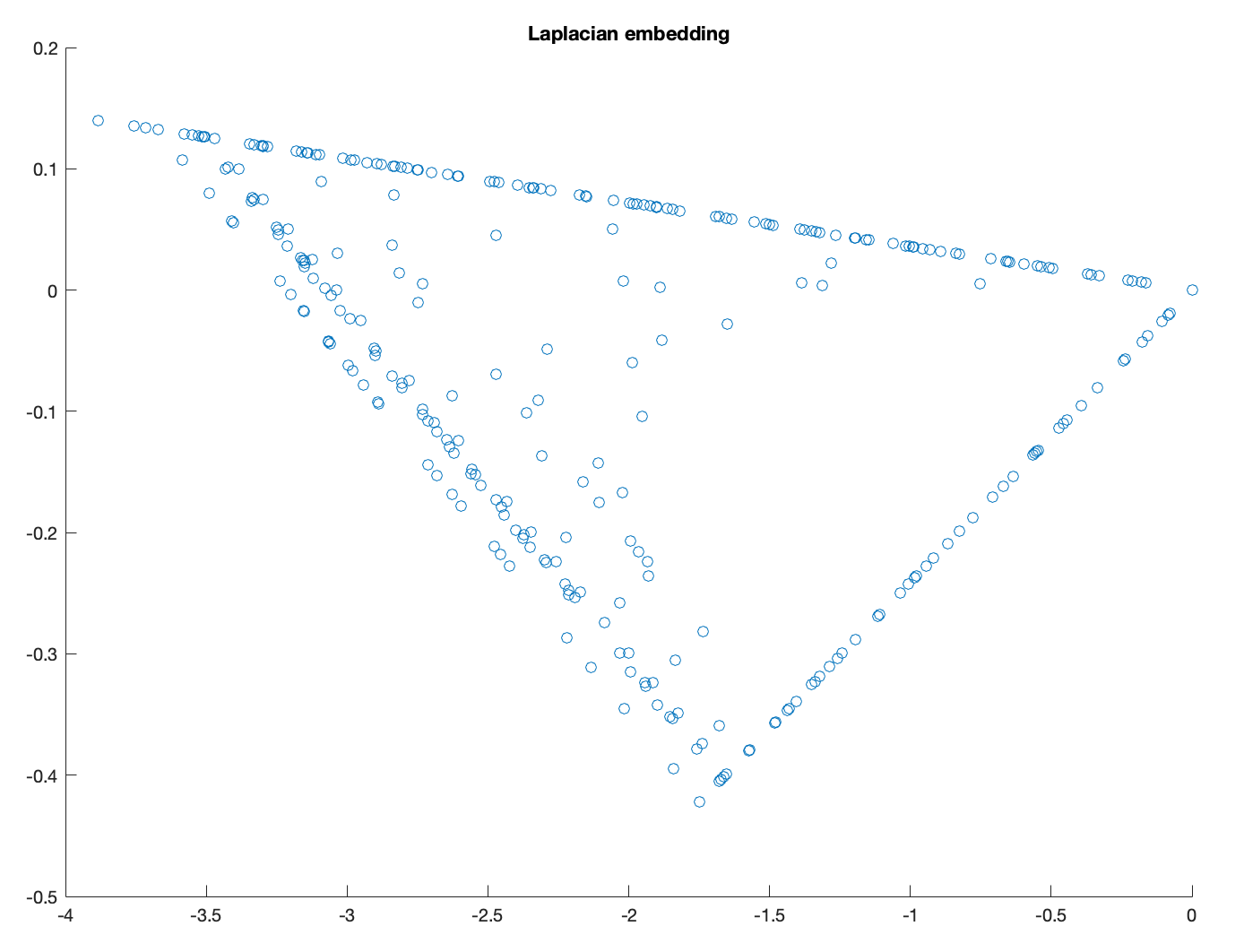}
	\caption{Laplacian embedding using all features}
\end{figure}

\section{High demand region identification}
As mentioned in previous section, the results presented for embedding were through graph convolution network (GCN) with foward pass only. There was no back propagation and the weights were not tuned. Therefore, the embeddings had a huge overlap of nodes in similar locations. In order to train the GCNnetwork and further identify which nodes have a traffic overload given traffic data of some other nodes, the labels $y$ are created for all nodes,

\begin{equation}
	y_x=\begin{cases}
		1, & \text{if $x_f>=\kappa$}.\\
		0, & \text{otherwise}.
	\end{cases}
\end{equation}

Where the node is represented by $x$ and its traffic value by $x_f$. The traffic is compared against a pre-determined threshold $\kappa$. I chose this $\kappa$ such that a fair amount of instances from both classes exist in the dataset.
The objective of this section is to do both embedding and estimation of node labels. Thus, in addition to GCN we utiize fully connected softmax activation at end of network. furthermore, we utilize cross entropy loss function  for backpropagation and learning the model weights for binary classification,

\begin{equation}
	Loss = -yllog(p) + (1-y)log(1-p)
\end{equation}
Where y is indicator whether the predicted label is correct or not and p is probability of class label 1.  Please note that the loss function is computed only using the nodes whose labels are available. The architecture for the overall network is shown below. A and X represent adjacency matrix and feature matrix respectively. The size of input network was huge, therefore a dropout layer was used to reduce the density of connections.

\begin{figure}[h!]
	\includegraphics[scale=0.65]{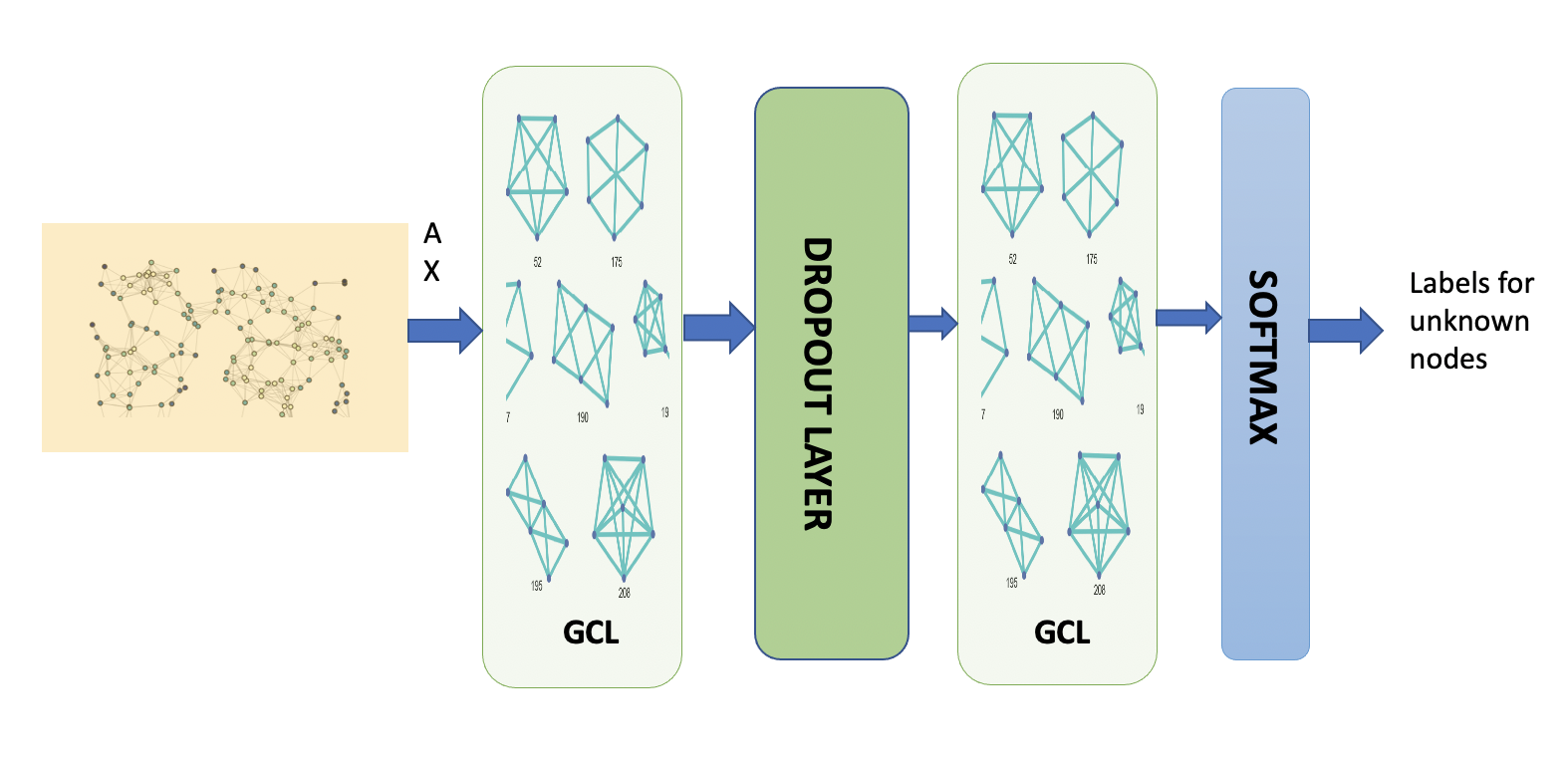}
	\caption{Graph NN architecture for learning }
\end{figure}

\section{Spatiol-Temporal  Traffic Prediction}
As mentioned in the introduction, cellular traffic is highly dynamic entity. The demands changes over both and short periods. However, there some level of periodicity or repetition in pattern resulting from repetition in daily routines of people living in certain area. In order to accurately characterize spatial heat map of traffic data, it is critical to utilize the temporal data as well. Please note that time is just not any other dimension or adding time to 2D space is not equivalent to a 3D system. A dedicated solution is needed to deal with time.  \\

In previous works, recurrent neural networks (RNN)have been used with LSTM where LSTM acts like a memory component and provides a track of past data to utilize in current estimation and future prediction\cite{time1}.  Another work \cite{time3} utilizes two RNN networks, one for nodes and one for edges to take care of both spatial and temporal aspects. The are several other works based on RNN that deal with spatial-temporal data. However, RNNs follow $y=wx+bias$ and require learning of weight matrix W whose size is equal to the input. They compute gradients for backpropagation through iterative methods which is time consuming and also suffer from gradient explosion/vanishing. Compared to RNN, Graph Convolution Neural networks (GCNN) do not depend on input size, are faster and more efficient in terms of computing cost and memory requirements.

In this part of project, I take time-stamped traffic data for different base stations $X\in R^{T\times N\times d} $, where $T$ is total number of time stamps, $N$ is number of nodes or Base stations and $d$ is number of features. The time stamps are uniformly distributed. At each time step $i$, the model observes the current and past data for $m$ time stamps : $X'=X(i-m:i)$ and predicts the traffic pattern for the next $k$ steps : $X^{pred} = \hat{X(i+1:i+n)}$.  

The NN model for predicting the traffic values is inspired from the work in \cite{time2}\cite{NN}\cite{main}. The authors have used chebychev polynomial for graph Convolution layer and recurrent gated unit to take care of time. Chebychev polynomial based embedding requires computation of eigenvalues and eigenvectors of Laplacian which is an expensive operation for a bog network like considered in this project. Therefore, I have used simple embedding based on adjacency matrix only which is described in previous section. Compared to using Recurrent unit to deal with temporal aspect, \cite{time2} has used a 1D Convolution Neural Network which is computationally much cheaper. For the same reason, i have also used CNN. The architecture of spatial-temporal traffic prediction model is presented in the figure 7.

\begin{figure}[t!]
	\centering
	\includegraphics[scale=0.63]{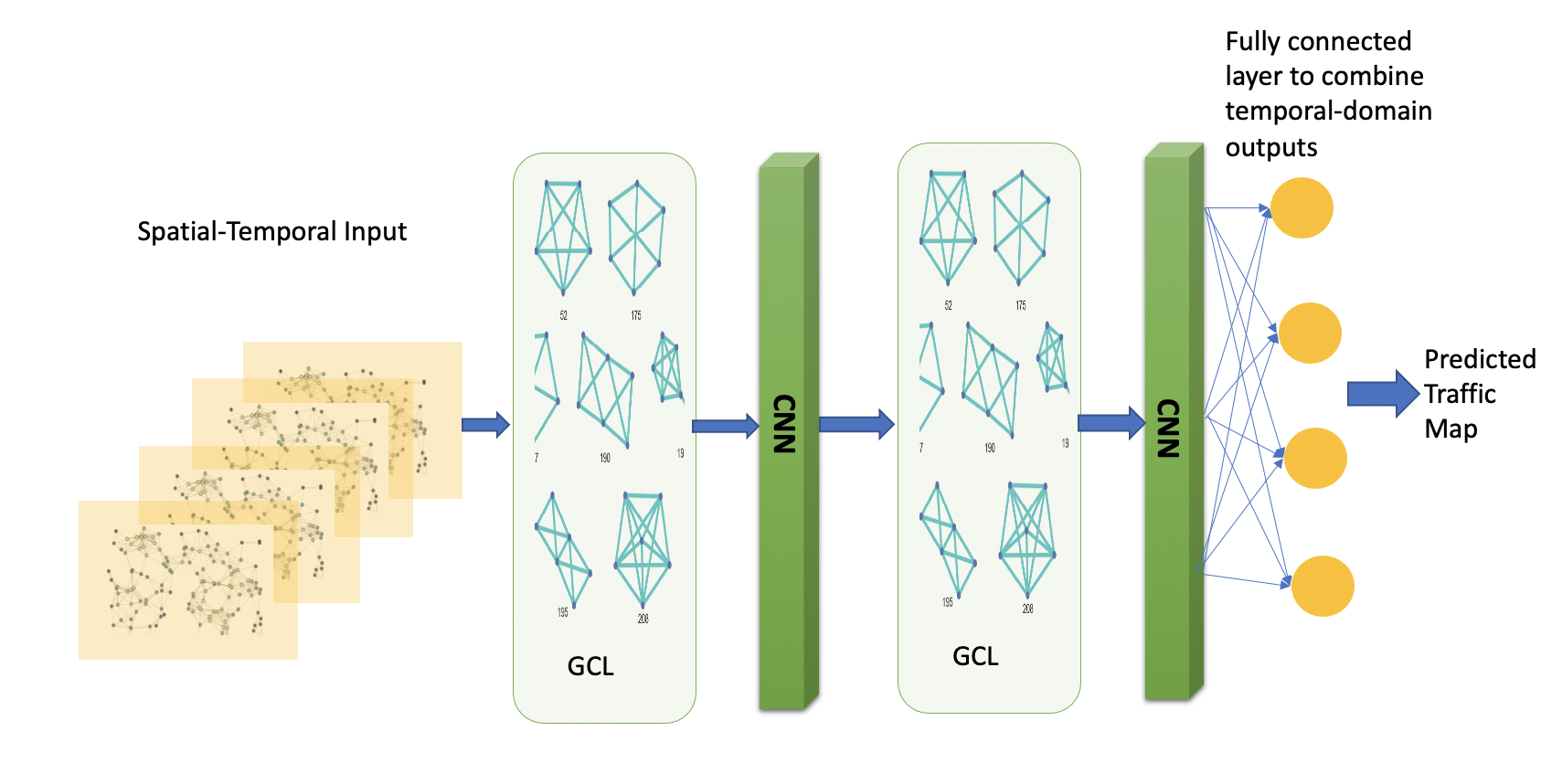}
	\caption{Spatial-Temporal Graph NN architecture }
\end{figure}

\section{Implementation and Results}
\begin{figure}[t!]
	\centering
	\includegraphics[scale=0.6]{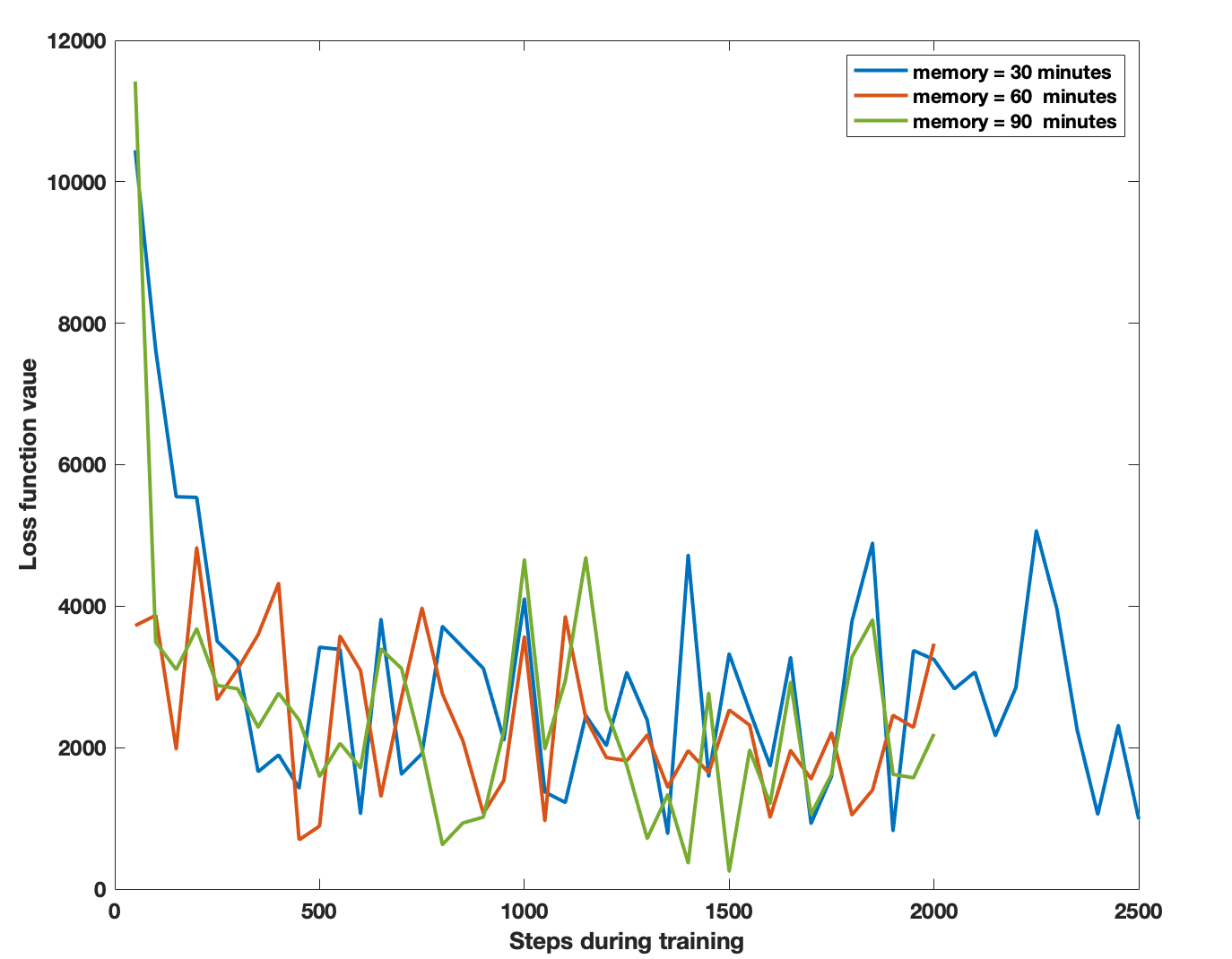}
	\caption{Loss function evolution during training }
	\label{m3}
\end{figure}
The dataset was preprocessed in MATLAB first. The Milano city contained 10000 nodes. However, even with 3000 nodes, the neural network training and testing was either taking too long or the IDE was crashing. Therefore, I created a small dataset of 1000 nodes to test and validate the implementation.  The dataset was provided in form of time-stamped base station traffic values. I created snapshots 10 minutes apart for multiple days. There were missing values too that I replaced with zero to avoid getting NaN later in the code.

For testing the embedding, I used MATLAB too and the results were provided in previous sections. However, for Graph convolution e network, I had to implement graph conv layer in Matlab as it didn't had it in its machine learning toolbox. My GCN implementation in MATLAB worked fine in forward pass but gave errors in backward propagation when the training module was feeding it with different input sizes. I could not resolve this issue so I decided to import graph convolution layers from Keras. However, MATLAB supports only certain versions of python and I was unable to find the open source code of compatible version of conv layer in Keras. To counter this, I decided to store the preprocessed data in mat files and import them in Python. During my implementation, I had to fix a huge number of bugs resulting from compatibility issues between different versions of tensorflow,keras and python. After many attempts, finally the code worked when I downgraded all three to older versions. Now the neural network is implemented in python and data processing files are in MATLAB. In addition to using Keras and some other libraries, I also took help from opensource libraries available online.

For the Spatial-temporal model, the input data was divided into 144 slots per day and data for a month was stored in X. The loss function was L2 norm of error between predicted and true values and ADAM optimzer was used to learn the weights. For performance metric, the root mean squared error and average absolute error was used.

As mentioned in the previous section, the model observes data from current and past $m$ snapshots to predict `the next $k$ snapshots where each snapshot is 10 minutes apart. For $k=3$ and different values of $m$, the training process in shown in figure \ref{m3}.

\begin{figure}
	\centering
	\includegraphics[scale=0.6]{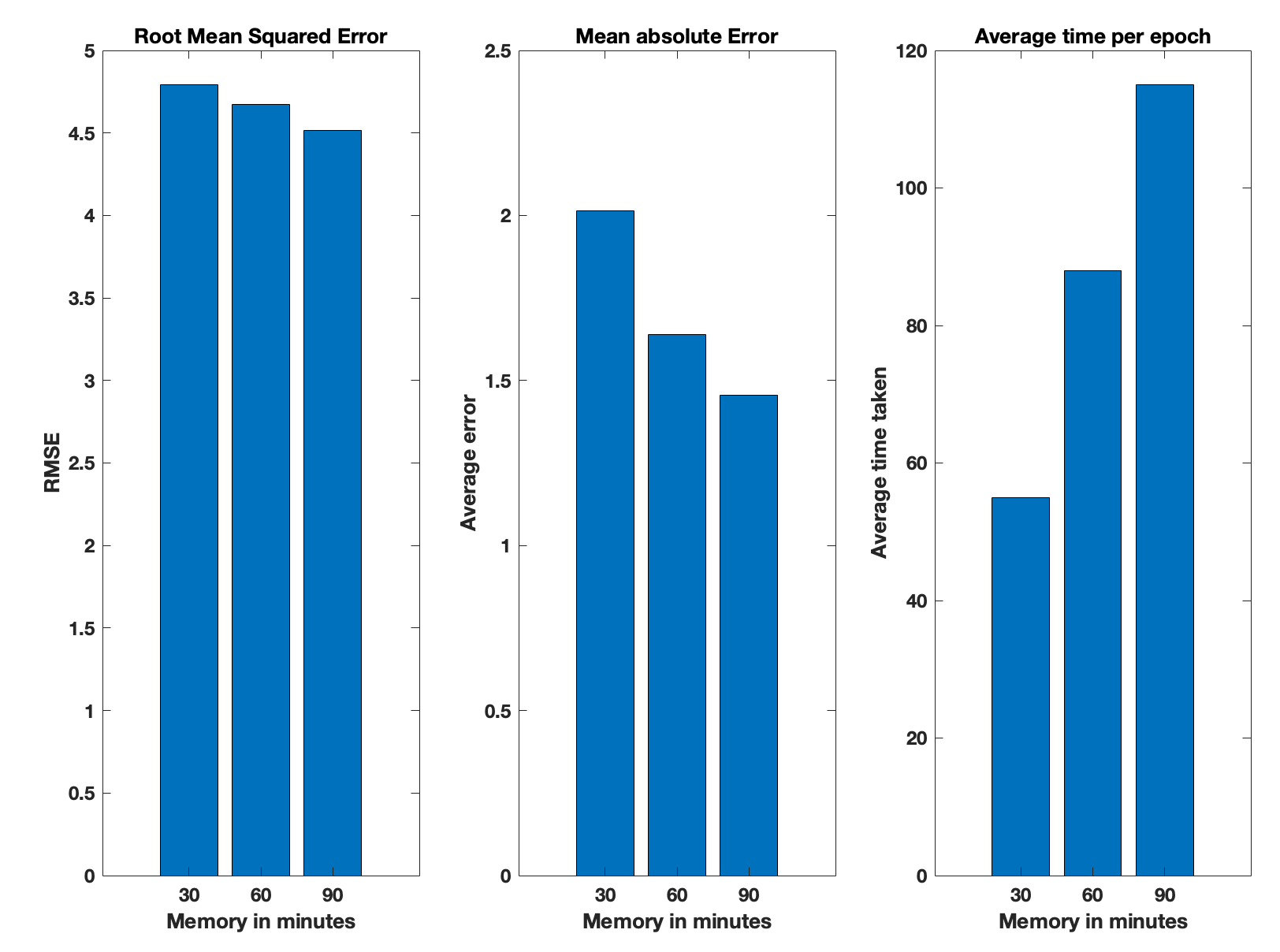}
	\caption{Test error performance for different $m$ }
	\label{m4}
\end{figure}

It can be observed that the loss function is fluctuating up and down. This is because of the fact that it goes up when new temporal data arrives and once it updates its parameters accordingly the error goes down. Furthermore, we can observe from the few starting steps that higher memory leads to loss going down faster which is intuitive as it can better learn the trend with more data. A more clear comparison between different memory sizes is provided in figure \ref{m4}.

As the model sees data for a longer past, the root mean squared error and average absolute error both go down. However, it not only comes at a cost of higher memory requirement but also higher time taken per epoch during training.

\begin{figure}[t!]
	\centering
	\includegraphics[scale=0.6]{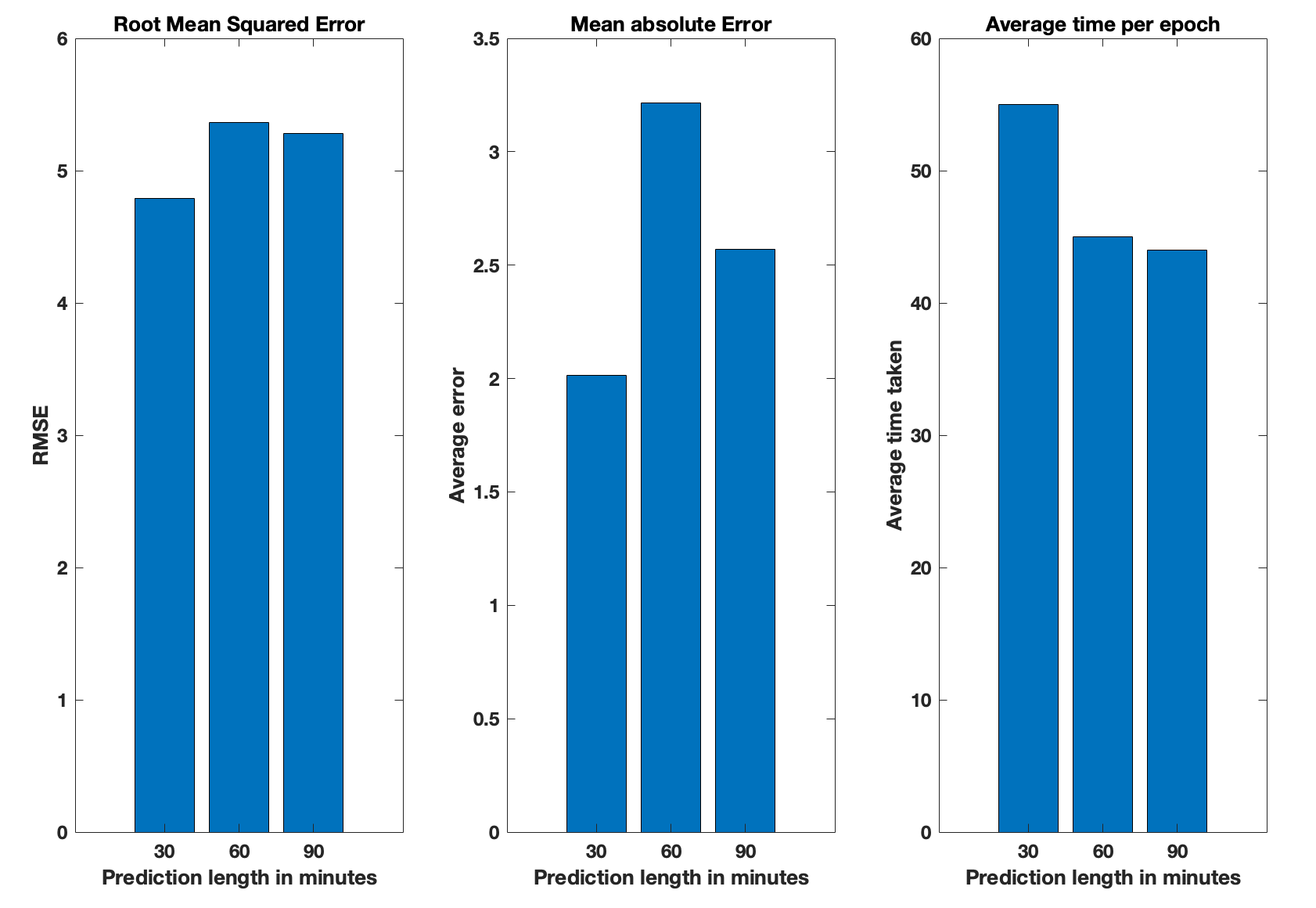}
	\caption{Test error performance for different $k$ }
	\label{n4}
\end{figure}

Next we keep the memory length fixed to 30 minutes but change the length of data that is predicted to observe how far the model can forecast into future. The results are presented in figure \ref{n4}. Initially we see a significant increase in error when prediction length is increased from 30 to 60 which is expected. However, the bar at 90 minutes is unexpected. The RMSE is almost same and average error has gone down. There could be two possible reasons for this: either the periodic trend that is observed at 30 minutes is repeated after 1 hour and therefore the overall error has gone down or the traffic values in 60-90 min interval are really small and the averaging has made the overall error smaller. For the training time, it is almost constant because larger prediction length only increases computational time for the loss function which is already really cheap to compute.

\section{Future Extensions}
There are a lot of possibilities to define more informative graphs on this traffic data. For instance a graph can be defined by generating the average cellular traffic load over a week  in each grid and if the demand values between nodes are close to each other they are connected.
Another possibility is representing the traffic values as weights of edges. A future extension could also be a NN-based model that is provided the snapshot data and it learns the weighted adjacency matrix A from it. \\
Another key issue with problems of this magnitude is the spatial extent of data. The number of nodes are so huge that even doing simple tasks is not possible. This problem can be dealt by utilizing clustering in such a way that the size is reduced but the underlying graph structure is also intact.\\
One major limitation of this project was that I used fixed adjacency matrix. Only the input feature values were varying in time. This is not the case in real-life and temporal dependence in graph structure also requires new NN architecture design which is a promising future direction.\\

\section{Conclusion}
The results presented in this work motivate the use of graph neural networks (GNN) in cellular traffic estimation and prediction. Compared to evaluating time series data using correlation or similar signal processing techniques, GNNs explore the underlying relationships between different base stations and leverage them to obtain more intelligent predictions. I have not presented the result for high demad region identification because of space constraints, but with only two layers the training and test accuracy was around 80-90\%.  GNNs also stand out from other ML methods such as recurrent neural networks when it comes to big datasets like the one considered here. They are faster and more efficient. Once a model has been built and trained it can be deployed in real-life setting. However, the main challenge lies in design of the model and data preprocessing. Most real-life datasets do not come in form of graph and deriving meaningful graphs from them is non-trivial. Testing whether one graph formulation is better than the other require the model that can output prediction and estimation accuracy whose design is itself not completely agnostic of graph structure. Thus, one has to go back and forth to find the optimum combination.

\bibliographystyle{IEEEtran}
\bibliography{Related}

\end{document}